%% file: main.tex
\documentclass[runningheads]{llncs}

\usepackage{amsmath,amsfonts,amssymb,amsthm}
\usepackage{verbatim}
\usepackage{color}
\usepackage{epsfig}
\usepackage{array}
\usepackage{algorithm}
\usepackage{cite}
\usepackage{tabu}
\usepackage{mathtools}

\usepackage{wrapfig}
\usepackage{MyMnSymbol}
\def\etal{\emph{et al. }}

\begin{document}

\title{Placental Flattening via Volumetric Parameterization}
\titlerunning{Placental Flattening via Volumetric Parameterization}  
%
\author{S. Mazdak Abulnaga\inst{1}\and Esra Abaci Turk\inst{2} \and Mikhail Bessmeltsev\inst{3} \and P. Ellen Grant\inst{2} \and Justin Solomon\inst{1} \and Polina Golland\inst{1}}
\index{Abulnaga, S. Mazdak}
\index{Abaci Turk, Esra} 
\index{Bessmeltsev, Mikhail}
\index{Grant, P. Ellen}
\index{Solomon, Justin}
\index{Golland, Polina}
\authorrunning{S. M. Abulnaga \etal} 
\institute{Computer Science and Artificial Intelligence Lab, MIT, Cambridge, MA, USA\\    
\email{abulnaga@mit.edu}\\  
\and Fetal-Neonatal Neuroimaging and Developmental Science Center, Boston Children's Hospital, Harvard Medical School, Boston, MA, USA\\
\and Department of Computer Science and Operations Research, Universit\'e de Montr\'eal, Montr\'eal, QC, Canada
}

\maketitle
\begin{abstract}
\input{sections/0-abstract}
\end{abstract}
\input{sections/1-intro}

\input{sections/2-methods}
\input{sections/3-results}
\input{sections/4-conclusion}

\bibliographystyle{splncs04}
\bibliography{main}

\end{document}

%% file: sections/0-abstract.tex
We present a volumetric mesh-based algorithm for flattening the placenta
to a canonical template to enable effective visualization of local
anatomy and function. Monitoring placental function \textit{in vivo} promises
to support pregnancy assessment and to improve care outcomes. We aim
to alleviate visualization and interpretation challenges presented by
the shape of the placenta when it is attached to the curved uterine
wall. To do so, we flatten the volumetric mesh that captures placental shape to
resemble the well-studied \textit{ex vivo} shape. We formulate our
method as a map from the \textit{in vivo} shape to a flattened template that minimizes the symmetric Dirichlet energy
to control distortion throughout the volume. Local injectivity is enforced via constrained line search during gradient descent. We
evaluate the proposed method on 28 placenta shapes extracted from MRI images in a clinical study of placental function. We achieve sub-voxel accuracy in
mapping the boundary of the placenta to the template while
successfully controlling distortion throughout the volume. We illustrate how the resulting mapping of the placenta enhances visualization of placental anatomy and function. Our implementation is freely available at \url{https://github.com/mabulnaga/placenta-flattening}.

\keywords{Placenta, Fetal MRI, Flattening, Injective maps, Volumetric mesh parameterization, Anatomy visualization }

%% file: sections/1-intro.tex
\section{Introduction}
The placenta is a critical organ that
connects the fetus to the maternal blood
system. Placental dysfunction increases
the risk of pregnancy complications, with long-term effects on a child's health
and development and the mother's health. It is therefore critical to monitor placental
function and health \textit{in vivo}. Ultrasound and MRI capture
detailed information about the placental position, shape, and tissue
properties~\cite{sorensen2013changes}. Blood 
oxygen level dependent (BOLD) MRI has recently been demonstrated to
assess oxygen transport within the
placenta~\cite{sorensen2013changes,luo2017vivo}, providing initial
evidence for clinical utility of MRI for functional assessment of the placenta. 
Compared with ultrasound, MRI provides direct measurements of placental function, which provides signals necessary to study function and assess pathology~\cite{sorensen2013changes,luo2017vivo}. 

The \textit{in vivo} shape of the placenta is determined by
the curved surface of the uterine wall to which it is attached during
pregnancy. This presents significant challenges for interpretation of
the MRI scans. No common standard exists for visualizing the functional or anatomical images of the placenta whose \textit{in vivo} shape and location of attachment to the uterine wall vary greatly across subjects. We present a novel algorithm for mapping the placental shape observed in an MRI scan to a flattened template that resembles the organ's well-studied \textit{ex vivo} flattened shape, to alleviate visualization challenges during \textit{in vivo} examination and to facilitate clinical research and development of placental health biomarkers. Our work offers the first step toward developing a common coordinate system to enable statistical analysis. 

We build on state-of-the-art mesh parameterization methods to
represent and estimate the deformation of the placenta onto a
template. Mesh parameterization is a topic of active research in
geometry processing for mapping surfaces to canonical domains such as
planes or spheres while guaranteeing desirable
properties of the mapping such as
injectivity~\cite{rabinovich2016scalable,smith2015bijective}. When
applied to cortical mapping, this parameterization facilitates
visualization and population studies~\cite{fischl1999cortical,timsari2000optimization,tosun2001hemispherical}. The formulation may estimate the optimal map by minimizing a cost function that penalizes areal
distortion~\cite{tosun2001hemispherical}, changes in geodesic
distance~\cite{fischl1999cortical}, or more commonly a combination of
different distortion measures~\cite{timsari2000optimization}. In contrast to the inherently two-dimensional (2D) cortex however, the placenta is a fully three-dimensional (3D) organ. Image information along the depth direction from the maternal to the fetal side of the placenta is important for characterizing function. The parameterization must therefore map the entire volume.

In placenta imaging, a method for placental flattening has been
recently demonstrated~\cite{miao2017placenta}. Their algorithm
represents the placenta as a stack of parallel surfaces spanning the
thickness of the organ. Each surface is flattened separately by mapping the boundary to a disk with each interior vertex moved to the average of its neighbors. This approach is limited in two ways. First, the lack of correspondence across surfaces throughout the volume results in through-plane artifacts, distorting important depth-wise information. Second, due to the variability in placental shape, constraining the mapping to a fixed boundary results in high distortion. In contrast, our method computes a continuous volumetric mapping with free boundary that ensures uniform consistency throughout the volume and enables explicit control of the resulting deformation.

To our knowledge, this is the first volumetric approach for mapping the placenta to a canonical domain. Our algorithm estimates the transformation of the volumetric mesh to a template as the solution of an optimization problem. This formulation readily accepts a broad family
of templates and shape distortion functions. We choose to minimize the symmetric Dirichlet
energy~\cite{schreiner2004inter,smith2015bijective} to penalize local deformations of the 
volumetric mesh. We evaluate our method on images from a clinical MRI study, demonstrating effective mapping of the highly variable placental shape to the template with minimal distortion. We demonstrate improved visualization of anatomical structures and their surrounding context, illustrating the promise of our algorithm to support clinical use of MRI in placental imaging.

%% file: sections/2-methods.tex
\section{Methods}

We represent the placental shape as a tetrahedral mesh that
contains $N$~vertices and $K$~tetrahedra. In our experiments, we extract such meshes from segmented MRI scans as described later. We parameterize 
the mapping via mesh vertex locations in the template coordinate
system and interpolate the
deformation to the interior of each tetrahedron using a locally affine (piecewise-linear) model. The mapping is geometry-based so it is independent of imaging modality used.

\subsection{Problem Formulation}
Let $X \in \mathbb{R}^{3\times N}$ be a matrix whose columns are the 3D coordinates of all mesh vertices
in the template space with the $M$ boundary vertices forming the first
$M$ columns. Let
$X_k \in \mathbb{R}^{3\times 4}$ be a matrix whose columns are the 3D
coordinates of the four corner vertices of tetrahedron $k$ ($k =
1,\ldots, K$) in the template space. We formulate the mapping as an optimization problem over the mesh vertices that seeks to map to the template space while minimizing shape distortion. We formulate the general objective function as
\begin{equation}
\label{eqn:objectivefun}
\phi(X)= 
\underbrace{\sum_{m=1}^M A_m T\left(x_m\right)}_{\text{Template match}}  + \underbrace{\lambda \sum_{k=1}^K V_k
\mathcal{D}\left(X_k\right)}_{\text{Distortion}},
\end{equation}
where $\{x_m\}_{m=1}^{M}$ are the boundary vertex coordinates in the template system, $T\left(\cdot\right)$ is a measure of distance from the template shape, $A_m$ is the normalized barycentric area of boundary vertex $m$ in the original space, $\mathcal{D}\left(\cdot\right)$ measures local distortion, $V_k$ is the normalized volume of tetrahedron $k$ in the original space, and $\lambda$ is a parameter that governs the trade-off between the template fit and the shape distortion. The distortion term regularizes the mapping.

We use a locally affine model to capture the deformation of tetrahedron $Z_k$ in the original image space to $X_k$ in the template space. The Jacobian matrix $J(X_k) = \left(X_k B \right) \left(Z_k B \right)^{-1}$ captures the linear transformation of the new vertex coordinates $X_k$ while ignoring the shared translation component. The constant matrix $B \in \mathbb{R}^{4 \times 3}$ extracts three basis
vectors defining the tetrahedron.

We measure
local distortion using the symmetric Dirichlet energy density
\begin{equation}
\label{eqn:dirichletenergy}
\mathcal{D}(J) = \|J\|_{F}^{2} + \|J^{-1}\|_{F}^{2} = \sum_{i=1}^{3}\left(\sigma_{i}^2 + \sigma_{i}^{-2}\right),
\end{equation}
where $\| \cdot \|_{F}$ is the Frobenius norm and $\{\sigma_{1},
\sigma_{2}, \sigma_{3}\}$ are the singular values of matrix $J$~\cite{rabinovich2016scalable,smith2015bijective}.
We chose the symmetric Dirichlet energy since it penalizes expansion and shrinking equally and favors a locally-injective mapping.

The image intensities are mapped to the template space using barycentric coordinates (BC). The BC represent a voxel in a tetrahedron as a convex combination of the tetrahedon's vertices. Since our mapping is injective and affine, the resulting voxel position is determined using its BC and the mapped vertices.

\subsection{Template}
After evaluating several volumetric templates (ellipsoid, sphere, cylinder; not shown), we find that the best results (i.e., clear anatomical structure and small local distortion) are achieved by encouraging constant height of the flattened placenta to mimic the post-delivery examination process, where the maternal side is placed on an examination table. The fetal side is flattened to the ease visualization. The function $T(\cdot)$ measures the distance to the appropriate plane:
\begin{equation}
\label{eqn-template-term}
T(x) =
\begin{cases}
(x^{(3)} - h)^2 & \text{if } x \in \mathcal{F}(\partial Z), \\
(x^{(3)} + h)^2 & \text{if } x \in \mathcal{M}(\partial Z), \\
0 & \text{otherwise},
\end{cases}
\end{equation}
where $x^{(3)}$ refers to the third coordinate of point $x$ in the template coordinate system, $\partial Z$ denotes the mesh boundary in the original image space, and $\mathcal{F}(\partial Z)$, $\mathcal{M}(\partial Z)$ denote the fetal and maternal sides of $\partial Z$. We identify the maternal and fetal sides via spectral clustering~\cite{ng2002spectral} as described below.  

We employ a similarity metric based on the angle between unit normals of boundary vertices ($\hat{n}$). We construct an affinity matrix $W \in \mathbb{R}^{M\times M}$ whose ($i,j$) element $w_{i,j} = \exp\left\{\gamma\left(\hat{n}_i^{\top}\hat{n}_j\right)\right\}$ for any two boundary vertices $i$ and $j$ that are connected by a path of at most $3$ edges (the $3$-ring neighborhood), and $w_{i,j}=0$ otherwise. The parameter $\gamma$ penalizes the variation in the orientation of the normals. We cluster the boundary vertices by thresholding the values of the second smallest eigenvector of the Laplacian $L = I - D^{-\frac{1}{2}}WD^{-\frac{1}{2}}$, 
where $D$ is a diagonal matrix with $d_{i,i} = \sum_j w_{i,j}$ and $I$ is the identity matrix~\cite{ng2002spectral}. Since the maternal side is more curved, we assign the corresponding label to the cluster with the larger number of vertices on the convex hull of the mesh.

We use the term \emph{rim} to denote the highly curved region that separates the fetal and maternal sides. We first assign to the rim all vertices on the boundary of the two clusters, i.e.,\ those with neighbors in the other cluster. The rim is then dilated to a set width based on the approximated geodesic distance along the mesh boundary. The geodesic distance accounts for mesh irregularities and ensures a consistent rim width. 

\subsection{Optimization}
\label{s:optimization}
We minimize cost function $\phi(\cdot)$ in~\eqref{eqn:objectivefun} using gradient
descent. We initialize the mapping using the identity transformation. The gradient of the template term is linear in the vertices, $\frac{\partial T(x)}{\partial x^{(3)}} = 2\left(x^{(3)} \pm h\right)$. We derive the gradient of the symmetric Dirichlet energy term defined in~\eqref{eqn:dirichletenergy} using
the chain rule for matrices:
\begin{align}
\frac{d\, \|J(X_k) \|_F^2}{d\, X_k} & = 
X_k \left[ 2 B \left(Z_{k} B\right)^{-1} \left(Z_{k} B \right)^{-T} B^T\right], \nonumber \\
\frac{d\, \|J(X_k) ^{-1}\|_F^2}{d\, X_k} &= -
 2 X_k B \left(B^T X_k^T X_k B \right)^{-1} B^T Z_k^T Z_k B \left(B^T X_k^T X_k B \right)^{-1} B^T. \nonumber
\end{align}

We employ line search to prevent tetrahedra from ``flipping,'' i.e.,
from crossing the singularity point of zero volume, thereby enforcing local injectivity~\cite{smith2015bijective}. In every iteration, we determine the largest value $\eta$ such that adjusting
the current vertex locations~$X$ by $- \eta \nabla \phi(X)$
avoids singularities for all tetrahedra. The (signed) volume of
tetrahedron $k$ is computed as the determinant of matrix $\left(X_k-\eta \nabla \phi\left(X_k\right) B \right)$ and is a cubic polynomial of $\eta$. The smallest, positive real root provides the upper limit for $\eta$ in the line search~\cite{smith2015bijective}. 

\subsection{Implementation Details}

We generate tetrahedral meshes from segmentation labelmaps using
iso2mesh\cite{fang2009tetrahedral}. Prior to mapping, we center
the mesh and rotate it to align its principal axes with the template. We assume the algorithm convergences when the Frobenius norm of the gradient is lower than $1 \times 10^{-4}$. We implemented the algorithm in MATLAB using GPU functionality to
parallelize computation, and ran our experiments on an NVIDIA Titan Xp (12GB) GPU. The algorithm took an average of $4083$ iterations and took less than $20$ minutes to converge. Our implementation is freely available at \url{https://github.com/mabulnaga/placenta-flattening}.

%% file: sections/3-results.tex
\section{Experiments}
\label{s:experiments}

{\bf Data: } We validate the approach on a set of 28 MRI scans from 2 studies.
The first is a twin study on 7 pregnant women (gestational age (GA):
28--34 weeks). All twin pregnancies had one placenta shared by the 2 twins. The second is a singleton pregnancy study on 11 women (GA: 27 -- 40 weeks). For 10 of 11 subjects, scans were acquired in the supine and left lateral positions, yielding 20 different segmentations. MRI BOLD scans were acquired on a 3T Siemens Skyra scanner (GRE-EPI, 3 $\text{mm}^3$, TR=5.8--8s, TE=32--36ms, FA= 90$^\circ$). The placenta was manually
segmented by a trained observer and input to the meshing software, which produced 6,500 tetrahedra and 2,800 surface
triangles on average.

\begin{figure}[!t]
\centering
\includegraphics[width=0.8\textwidth]{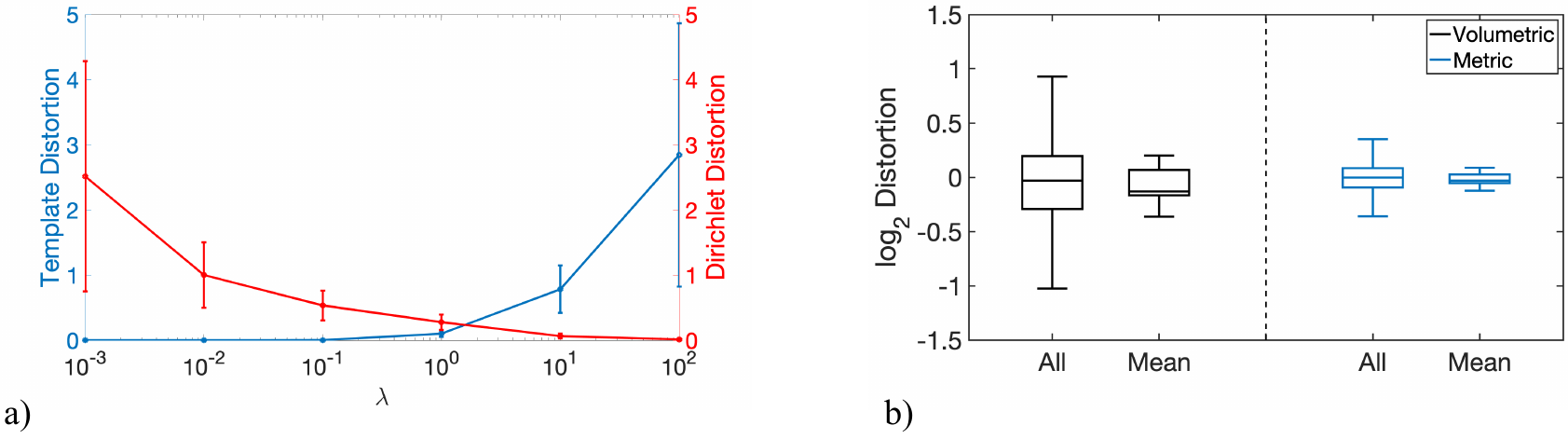}
\caption{ a) The final template matching term and distortion energy approaching optimality near $\lambda=1$; b) Distributions of distortion. We report (i) the statistics of volumetric and metric distortion across all tetrahedra (All), weighted by original tetrahedral volume, and (ii) of the mean distortion values across the 28 cases (Mean). The distributions are unimodal and well-contained at the extrema.}
\label{f:param-distortion}
\end{figure}

\noindent {\bf Parameters: } We used a grid search to determine the values of the hyper-parameters. We set shape distortion parameter $\lambda = 1$ as it was in the optimal trade-off range between the template match and distortion (Fig.~\ref{f:param-distortion}a). We set the template half-height $h$
to be half of the placenta thickness estimated from the histogram of
the distance transform values inside the segmentation boundary. We set the spectral clustering parameter $\gamma=20$ and used a boundary geodesic distance of $5$ voxels as the width of the rim.

\noindent {\bf Evaluation:} We use the log-determinant of the
Jacobian matrix $\log_2\det\left(J\left(X_k\right)\right)$ of tetrahedron $k$ to quantify local volumetric distortion~\cite{leow2007statistical}. We quantify metric distortion using the ratio of edge lengths $\log_2\left(x_{ij}/z_{ij}\right)$. We visually assess the quality of the
transformation by mapping the BOLD MRI to the template
coordinate system.

We compare with the prior parameterization approach in~\cite{miao2017placenta} where 2D surfaces spanning the placenta were independently parameterized to a disk in $\mathbb{R}^2$. The surfaces were derived by cutting Euclidean level sets to minimize local curvature changes. To emulate this method, we derive such 2D surfaces by intersecting the flattened placenta volume with planes spaced one voxel apart.  We harmonically parameterize each surface to a disk~\cite{joshi2007harmonic}, each with a corresponding point mapped to the north point of the disk. We scale the areas and edge lengths in the parameterized space to have mean $0$ areal and metric log-distortion.

\begin{figure}[!bt]
\centering
\includegraphics[width=0.8\textwidth]{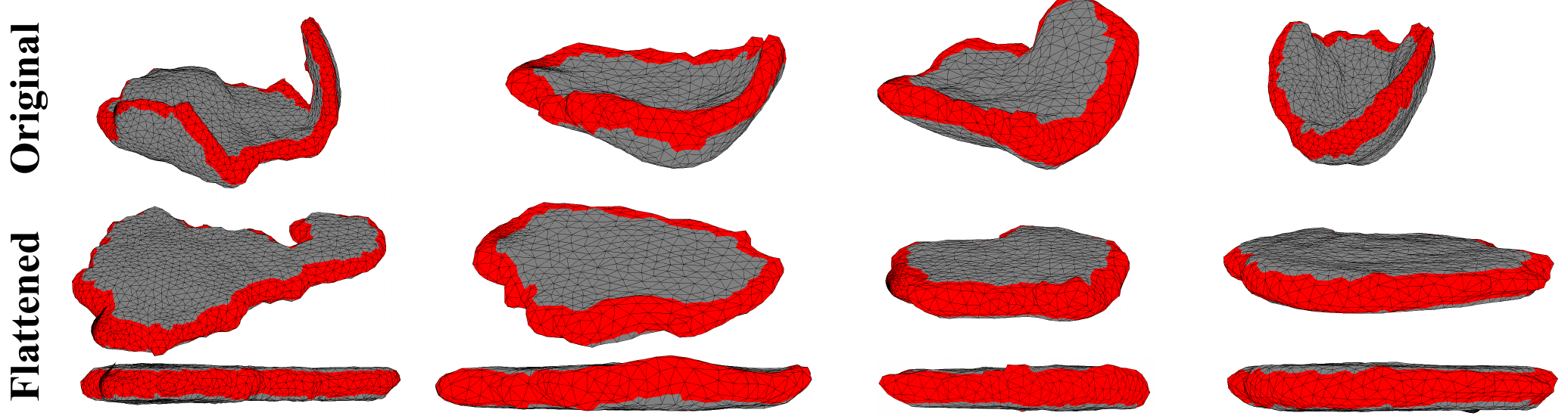}
\caption{Flattening results for four tetrahedral meshes: One twin (left) and three singleton pregnancies. Fetal sides are facing upwards and the rim is shown in red. The greatly differing shapes in the dataset are robustly mapped and the rim is a faithful representation of the curved area separating the fetal and maternal sides.} 
\label{f:4subjects-mesh}
\end{figure}

\subsection{Results}
For all cases, our algorithm achieves sub-voxel
accuracy of matching the template (median of $0.09$\, voxels, max. of
$0.30$\,voxels). 
The resulting transformations achieve close to minimum values of the symmetric
Dirichlet energy (median $3.87\%$\, higher, max. $9.04\%$ higher than the smallest possible value), where the energy is minimized by the identity transformation. Fig.~\ref{f:param-distortion}b demonstrates the mapping achieves minimal local volumetric and metric distortions. We did not observe differences  across twin and singleton pregnancies. 

Fig.~\ref{f:4subjects-mesh} illustrates the mapping results for four placentae, highlighting the variability in shape encountered in the dataset. We are able to map difficult structures robustly such as the fold in the first placenta and bowl-shape in the last. The estimated rim also effectively separates the maternal and fetal regions.

\begin{figure}[!t]
\centering
\includegraphics[width=0.9\textwidth]{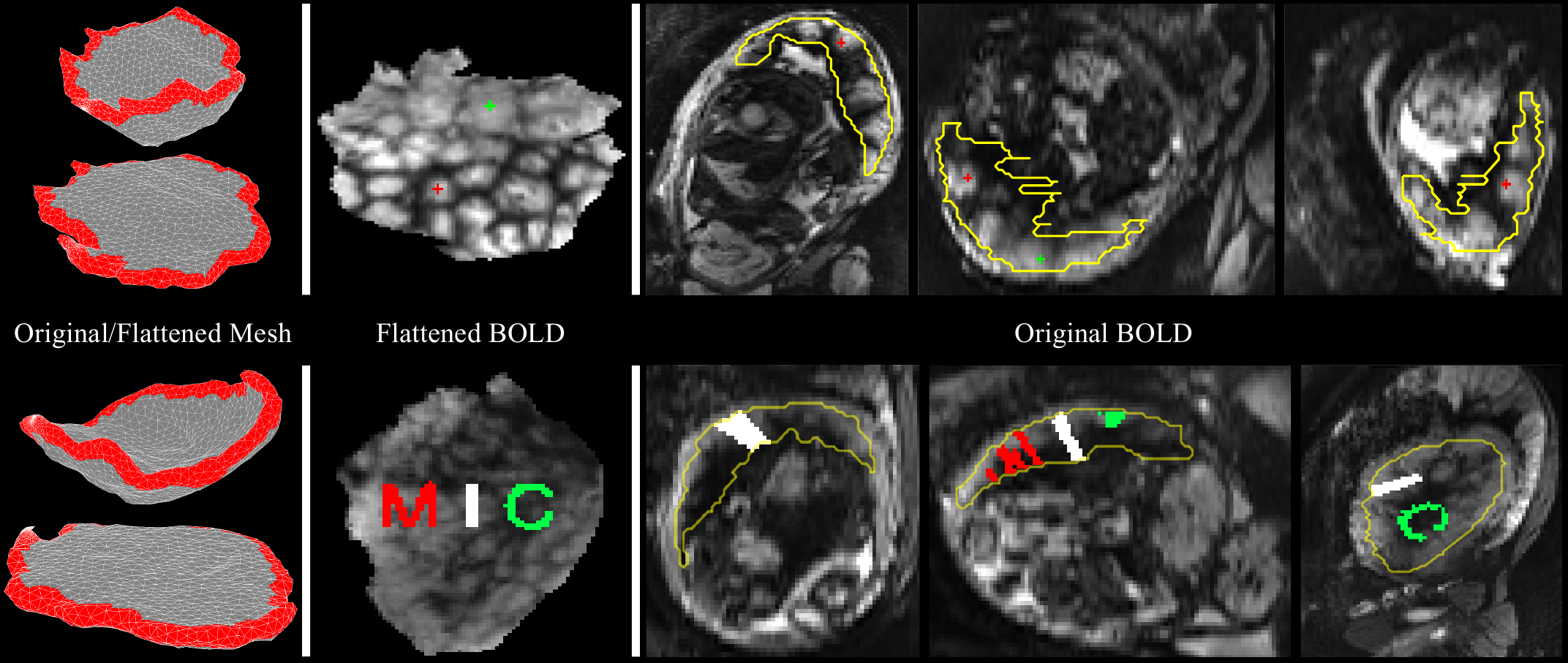}
\caption{Visual assessment of landmarks in two twin pregnancy subjects. Top: The cotyledons, characterized by a honeycomb structure of small hyperintense regions, are immediately apparent in the flattened view. The local anatomy surrounding the two cotyledons with marked centers is lost in the original images. 
Bottom: Mapping the fabricated landmark ``MIC" to the original volume loses contextual information since the curved geometry distorts the letters and they are not easily seen together. Local anatomy is difficult to visualize in the original volume due to the curvature of the uterine wall that determines the \textit{in vivo} shape of the placenta.}
\label{f:ipmi}
\end{figure}

Fig.~\ref{f:ipmi} demonstrates that the mapping enhances visualization of landmarks. The mapped BOLD intensity patterns clearly visualize local anatomy and function as is apparent in the honeycomb structure of the cotyledons, which are the circular structures that exchange oxygen and nutrients between the maternal blood and the fetal blood in the chorionic villi~\cite{luo2017vivo}. Cotyledons appear hyperintense in BOLD MRI. Similarly, contextual information is lost when visualizing a fabricated landmark (the letters ``MIC") in the curved volume. Since the inherent geometry of the placenta is flat, the letters could represent biomarkers. The curved geometry of the \textit{in vivo} placenta demonstrates immediate difficulty in visualizing the details that are clearly seen in the flattened view. Several views in the original image space are required to identify anatomical landmarks.

\begin{figure}[!t]
\centering
\includegraphics[width=0.9\textwidth]{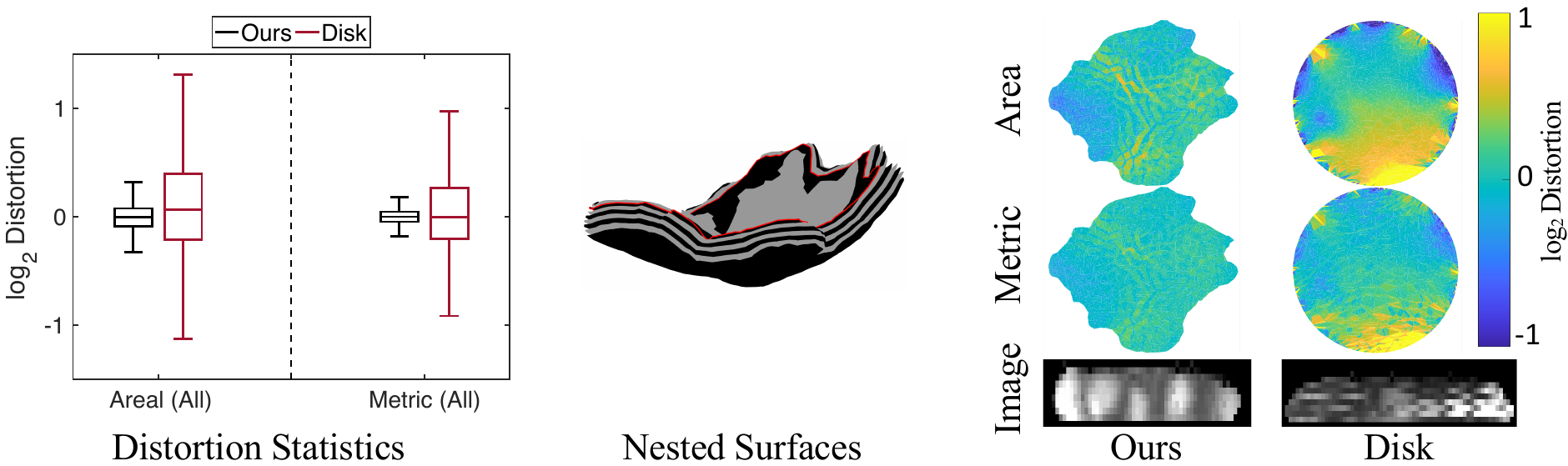}
\caption{Comparison with the baseline 2D parameterization approach. Left: Our method results in significantly lower distortion across all cases ($p<.001$). Right: Our free-boundary parameterization results in lower and more homogeneous spatial distributions of distortion as demonstrated on an interior surface. The baseline approach also creates through-plane imaging artifacts due to a lack of coupling across flattened planes.}
\label{f:param}
\end{figure}

Fig.~\ref{f:param} compares our method with the baseline 2D approach. Our method produces considerably lower distortion across all cases. Mapping to a fixed boundary results in higher distortion, and we observe image artifacts due to the lack of coupling across planes. These results confirm the need for a free boundary volumetric parameterization method. Finally, we note that the nested surfaces in the original space were derived from the volumetric parameterization to the flattened space. In practice, estimating these surfaces is challenging and requires a multi-step specialized pipeline as in~\cite{miao2017placenta}.

%% file: sections/4-conclusion.tex
\section{Conclusion}
We developed a volumetric mesh-based mapping of the placenta to a flattened template represented by two parallel planes, resembling the \textit{ex vivo} flattened shape of the organ, thereby enabling visualization of local anatomy and function. An immediate next step is to assess the utility for clinical research by quantifying the improvement in identifying known key biomarkers and anatomical features. In future work, we will improve the template using anatomical data such as the umbilical cord insertion. We are collecting higher-resolution anatomical images to include landmarks that are not easily seen in the BOLD (functional) images. 
This work is the first step towards developing a common coordinate system to visualize, examine, and study the organ as well as to support statistical analysis across subjects and time. Such a framework promises to advance the state of the art in studies of the placenta and to provide MRI biomarkers of fetal health.


\subsubsection*{Acknowledgments}
This work was supported in part by NIH NIBIB NAC
P41EB015902, NIH NICHD U01HD087211, NSF IIS-1838071, Air Force FA9550-19-1-0319, Wistron, SIP, AWS, NSF Graduate Research Fellowship, and NSERC Post Graduate Scholarship.